\documentclass{article}

\usepackage[nonatbib,final]{neurips_2020}
\usepackage[numbers]{natbib}

\usepackage[utf8]{inputenc} 
\usepackage[T1]{fontenc}    
\usepackage{hyperref}       
\usepackage{url}            
\usepackage{booktabs}       
\usepackage{amsfonts}       
\usepackage{nicefrac}       
\usepackage{microtype}      
\usepackage{xcolor}
\usepackage{xspace}
\usepackage{tabularx}
\usepackage{amsmath,bm,multicol,multirow}
\usepackage{cleveref}
\usepackage[pdftex]{graphicx}
\usepackage{array, makecell}
\usepackage[scientific-notation=true]{siunitx}
\usepackage{subcaption}
\usepackage{chngpage}
\usepackage{wrapfig,lipsum,booktabs}
\usepackage{pifont}
\usepackage{soul}
\usepackage{tikz}

\newcolumntype{L}{>{\raggedright\arraybackslash}X}

\usepackage{ulem}
\newcommand{\colormat}[7]{
\stackrel{#7}{
\vcenter{\hbox{ 
\begin{tikzpicture}[scale=0.3, align=center]
\filldraw[step=1,black,thin] (0.0, 0.0) grid (#1, #2);
\filldraw[shift={(#4, #3)}, fill=gray] (0, 0) rectangle (#6, #5);
\end{tikzpicture}
}}}}

\definecolor{prefix}{rgb}{0, 0, 0}
\definecolor{knob}  {rgb}{1, 0, 0}
\definecolor{bw}    {rgb}{1, 0, 0}
\definecolor{rw}    {rgb}{.7, .3, .3}
\definecolor{knob4}  {rgb}{0, .8, .2}
\definecolor{bw4}    {rgb}{0, .8, .2}
\definecolor{rw4}    {rgb}{.3, .7, .3}
\definecolor{knob3}  {rgb}{0.5, 0, 1}
\definecolor{bw3}    {rgb}{0.5, 0, 1}
\definecolor{rw3}    {rgb}{.5, .3, 1}
\definecolor{knob2}  {rgb}{0, 0, 1}
\definecolor{bw2}    {rgb}{0, 0, 1}
\definecolor{rw2}    {rgb}{0.3, .3, .9}

\definecolor{prefix}{rgb}{0, 0, 0}
\definecolor{knob}  {rgb}{0, 0, 1}
\definecolor{bw}    {rgb}{0, 0, 1}
\definecolor{rw}    {rgb}{.3, .3, .7}

\definecolor{lavenderblue}{rgb}{0.8, 0.8, 1.0}
\definecolor{amber}{rgb}{1.0, 0.75, 0.0}
\definecolor{azure)}{rgb}{0.0, 0.5, 1.0}
\definecolor{candyapplered}{rgb}{1.0, 0.03, 0.0}
\definecolor{emerald}{rgb}{0.31, 0.78, 0.47}
\definecolor{battleshipgrey}{rgb}{0.52, 0.52, 0.51}
\definecolor{green}{rgb}{0.0, 1, 0}
\definecolor{lavenderpurple}{rgb}{0.59, 0.48, 0.71}
\definecolor{ao}{rgb}{0.0, 0.5, 0.0}
\definecolor{eggplant}{rgb}{0.38, 0.25, 0.32}
\definecolor{dodgerblue}{rgb}{0.12, 0.56, 1.0}
\definecolor{deepmagenta}{rgb}{0.8, 0.0, 0.8}

\newcommand{\subtitle}[1]{\textcolor{prefix}{\textbf{[#1]}~}}
\newcommand{\sub}[1]{\textcolor{amber}{#1}}

\newcommand{\del}[1]{\textcolor{candyapplered}{#1}}
\newcommand{\correct}[1]{\textcolor{ao}{#1}}
\newcommand{\word}[1]{\textcolor{prefix}{\underline{#1}}}
\newcommand{\ignore}[1]{\textcolor{eggplant}{#1}}

\newcommand{\bw}[1]{\textcolor{bw}{#1}}   

\newcommand{\rwtwo}[1]{\textcolor{deepmagenta}{#1}}   

\newcommand{\cmark}{\ding{51}}%
\newcommand{\xmark}{\ding{55}}%

\title{\LARGE Towards Semi-Supervised\\ Semantics Understanding from Speech}

\author{%
    Cheng-I Lai\thanks{Work performed during an internship at Amazon AI.}\\
    MIT CSAIL\\
    \texttt{clai24@mit.edu} \\
    \And
    Jin Cao, Sravan Bodapati, Shang-Wen Li\thanks{Corresponding author.} \\
    Amazon AI\\
    {\tt \{jincao,sravanb,shangwel\}@amazon.com} \\
}

\begin{document}

\maketitle

\begin{abstract}
   Much recent work on Spoken Language Understanding (SLU) falls short in at least one of three ways: models were trained on oracle text input and neglected the Automatics Speech Recognition (ASR) outputs, models were trained to predict only intents without the slot values, or models were trained on a large amount of in-house data.
   We proposed a clean and general framework to learn semantics directly from speech with semi-supervision from transcribed speech to address these.
   Our framework is built upon pretrained end-to-end (E2E) ASR and self-supervised language models, such as BERT, and fine-tuned on a limited amount of target SLU corpus.
  In parallel, we identified two inadequate settings under which SLU models have been tested: noise-robustness and E2E semantics evaluation.
  We tested the proposed framework under realistic environmental noises and with a new metric, the slots edit $F_{1}$ score, on two public SLU corpora.
  Experiments show that our SLU framework with speech as input can perform on par with those with oracle text as input in semantics understanding, while environmental noises are present, and a limited amount of labeled semantics data is available.
\end{abstract}

\vspace{-4mm}
\section{Introduction}
\vspace{-2mm}

Spoken Language Understanding (SLU)\footnote{SLU typically consists of Automatic Speech Recognition (ASR) and Natural Language Understanding (NLU). ASR maps audio to text, and NLU maps text to semantics. Here, we are interested in learning a mapping directly from raw audio to semantics.} is at the frontend of many modern intelligent home devices, virtual assistants, and socialbots~\citep{yu2019gunrock,coucke2018snips}: given a spoken command, an SLU engine should extract relevant semantics\footnote{Semantics is commonly formulated as intent and slots in common benchmarking datasets like ATIS.} from spoken commands for the demanded downstream tasks.
Since the debut of the Airline Travel Information System (ATIS) project~\citep{hemphill1990atis}, the field has progressed from knowledge-based~\citep{ward1994recent,seneff1992tina,dowding1994gemini} to data-driven approaches, notably those based on neural networks. 
In the seminal paper on ATIS by Tur et al.~\citep{tur2010left}, incorporating linguistically motivated features for NLU and improving ASR noise robustness were underscored as the research emphasis for the coming years. 
Now, a decade later, the progress arose by self-supervised language models (LMs), such as BERT~\citep{devlin2018bert}, and E2E SLU~\citep{serdyuk2018towards,lugosch2019speech} seem to have responded to those problems posed in~\citep{tur2010left}. 
Nevertheless, we found the current research agenda lacks in several perspectives: model training when limited semantics labels are available, model robustness under realistic noisy environments, and model evaluation with E2E intent classification (IC) and slot labeling (SL) evaluation. 
In this paper, we proposed an SLU framework that (1) learns with limited semantics labels, (2) is end-to-end, and (3) is robust under environmental noises.
The framework consists of an ASR, and a masked language model pretrained on audio-text pairs without semantics labels and is evaluated with an E2E evaluation metric. 
We break our arguments down into two parts: Modeling and Evaluation (see Table~\ref{tab:comparison} for a comparison of our framework with previous work).

    
    \begin{figure}[!htbp]
    \centering
    \includegraphics[width=1.0\linewidth]{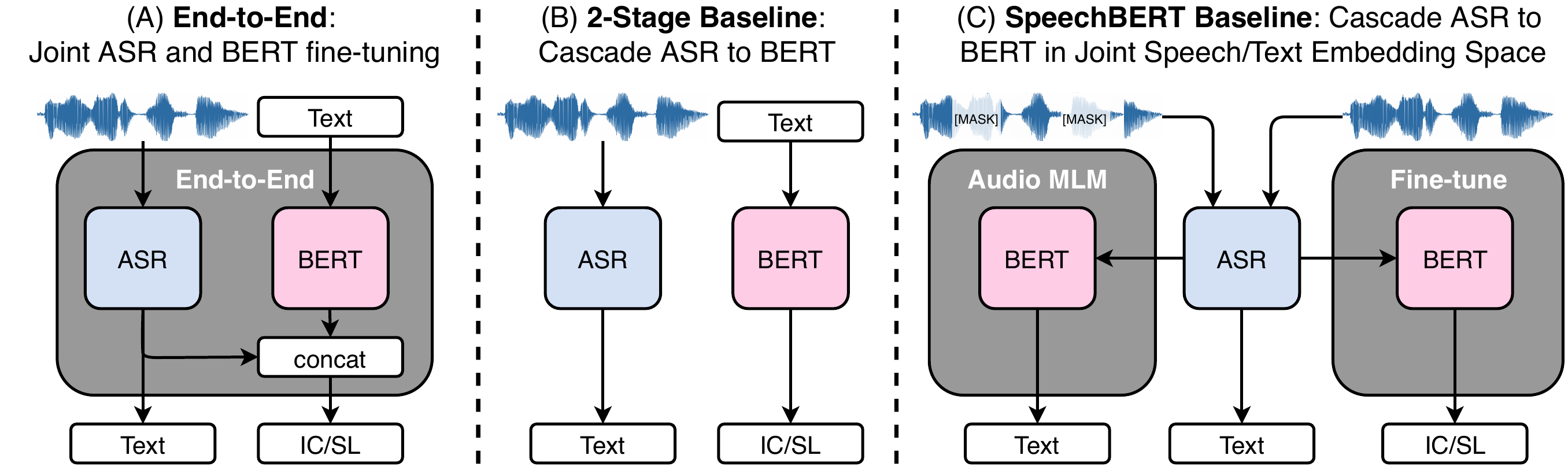}
    \caption{
    Our proposed semi-supervised E2E learning framework with ASR and BERT for joint intent classification (IC) and slot labeling (SL) directly from speech. 
    (A) shows the E2E approach, in which E2E ASR and BERT are trained jointly by predicting text and IC/SL. 
    (B) shows the 2-stage baseline, where text and IC/SL are obtained successively.
    (C) shows the SpeechBERT baseline, where BERT is adapted to take audio as input by first pretraining with Audio MLM loss and then fine-tuning for IC/SL. 
    A separate ASR is still needed for (B) and (C).}
    \label{fig:overview}
    \vspace{-6mm}
    \end{figure}

\paragraph{Why do we want semi-supervised learning for SLU?}
Neural networks benefit from large quantities of labeled training data, and one could train SLU models end-to-end with them~\citep{rao2020speech,coucke2018snips,haghani2018audio,serdyuk2018towards}. 
However, curating labeled IC/SL data is expensive, and often time only a limited amount of labels are disposable. 
Semi-supervised learning could be a useful scenario for training SLU models for various domains whereby model components are pretrained on large amounts of unlabeled data and then fine-tuned with target semantic labels. 
Despite some work already implemented this pretraining then fine-tuning scheme, they were limited such that (1) models require a separate ASR or some form of "feedback" from ASR, (2) models were designed to only predict intents without the slot values, or (3) models did not take advantage of the generalization capacity of self-supervised LMs, such as BERT, where we found to be essential to obtain competitive results if limited semantics labels are present. 
In contrast, our framework provides a clean and general solution to the above limitations. 
Our semi-supervised framework is a direct product of the self-supervised trend in speech and audio processing which takes the form of: predictive objectives~\citep{chung2019unsupervised,chung2020generative,chung2020improved,chung2020vector,pascual2019learning,chorowski2019unsupervised}, contrastive objectives~\citep{oord2018representation,kawakami2019unsupervised,baevski2020wav2vec,baevski2019vq,baevski2019effectiveness,schneider2019wav2vec,liu2020tera,chi2020audio,jiang2019improving,liu2020mockingjay,lai2019contrastive,kharitonov2020data}, grounded learnings~\citep{rouditchenko2020avlnet,khurana2020cstnet,conneau2020unsupervised,harwath2019learning,harwath2016unsupervised}, and self-trainings~\citep{hsu2020semi,xu2020iterative,chen2020semi,park2020improved,kahn2020self}.
Different from the above settings is that this work does not concern with learning general representations for several downstream tasks, nor does it rely on multiple modalities or pseudo labeling techniques. 
Our focus is on designing a better learning framework distinctly for \textit{semantics understanding} under limited labels.

\vspace{-3mm}
\paragraph{What's wrong with the current SLU evaluation setting?} 
Two significant bottlenecks of deploying SLU models into production are how prior work has evaluated them. 
First, SLU models were not trained and evaluated for noise-robustness. 
For example, benchmarking datasets ATIS and Fluent Speech Commands~\citep{lugosch2019speech} are very clean; conversely, SLU engines often operate under noises, such as environmental noises. 
Secondly, given that not until recently SLU has been composed of separately developed ASR and NLU components, there is little work on an E2E evaluation criterion.
Previous SL evaluation criterion does not consider a naturally occurring scenario where ASR hypothesis and human transcriptions have different lengths. 
Taking these into account, we proposed noise augmentation training for SLU and the slot edit $F_{1}$ score.
\vspace{-1mm}

Key contributions of this paper are summarized as follows:
\vspace{-1mm}
\begin{itemize}
  \item We introduced a semi-supervised framework for semantics understanding directly from speech to alleviate: 
  (1) the need for a large amount of in-house, homogenous data~\citep{rao2020speech,coucke2018snips,haghani2018audio,serdyuk2018towards} by pretrained components on transcribed speech 
  (2) the limitation of only intent classification~\citep{lugosch2019speech,huang2020learning,serdyuk2018towards} by predicting text, slots, and intents. 
  (3) any additional manipulation on labels or loss, such as label projection~\citep{caostyle}, output serialization~\citep{tomashenko2019recent,haghani2018audio,ghannay2018end}, ASR n-best hypothesis, or asr-robust training losses~\citep{huang2020learning,lee2019mitigating}. 
  Figure~\ref{fig:overview} illustrates our approach.
    
  \item The framework is trained with explicit noise augmentation such that it is robust to environmental noises and is evaluated with the slot edit $F_{1}$ score for end-to-end semantics evaluation.
  
  \item  Our framework improves upon previous work in Word Error Rate (WER) and IC/SL F1, and even rivaled its NLU counterpart with oracle text input~\citep{chen2019bert}. Experiments are conducted on public SLU corpora, ATIS, and SNIPS. We released the dataset used in this work. 
\end{itemize}




    
    
    \vspace{-5mm}
    \begin{table}[!htbp]
      \centering
      \caption{Comparison of our approaches with prior E2E SLU work. 
      Work indicated by a * formulated SL as an intent detection task (See Appendix~\ref{app:problem_with_FSC} for details). Full summary table is in Appendix~\ref{app:summary_table}.\\}
      \vspace{-3mm}
      \label{tab:comparison}
      \resizebox{\columnwidth}{!}{
      \begin{tabular}{lcccccc}
        \toprule
        \multirow{2}{*}{Model}  &  \multicolumn{4}{c}{Modeling}  &  \multicolumn{2}{c}{Evaluation}  \\
        \cmidrule(lr){2-5}\cmidrule(lr){6-7}
           &  Output  & Noise/Error & Semi-Supervised & E2E & E2E & Noise Robustness \\
        \midrule
        \multicolumn{7}{l}{\textbf{Proposed}}\\
        End-to-End  & text, intent, slots  &  \cmark & \cmark &  \cmark & \cmark & \cmark \\
        \midrule
        \multicolumn{7}{l}{\textbf{Our Baselines}}\\
        2-Stage     & text, intent, slots  &  \cmark & \cmark &  \xmark & \cmark & \cmark \\
        SpeechBERT  & text, intent, slots  &  \cmark & \cmark &  \xmark & \cmark & \cmark \\
        \midrule 
        \midrule
        \multicolumn{7}{l}{\textbf{Prior Work}}\\
        \citep{serdyuk2018towards}$^{*}$    & intent only   & \cmark & \xmark  & \cmark & \xmark & \cmark \\
        \citep{lugosch2019speech,wang2020large,cho2020speech}$^{*}$    &  intent only          & \xmark & \cmark  & \cmark & \xmark & \xmark \\
        \citep{radfar2020end}$^{*}$         & intent only   & \xmark & \xmark  & \cmark & \xmark & \xmark \\
        \citep{ghannay2018end}          & text, intent, slots   & \cmark & \cmark  & \cmark & \xmark & \xmark \\
        \citep{haghani2018audio}        & text, intent, slots   & \xmark & \xmark  & \cmark & \cmark & \xmark \\
        \citep{tomashenko2019recent}    & text, intent, slots   & \xmark & \cmark  & \cmark & \xmark & \xmark \\
        \citep{rao2020speech}           & text, intent, slots   & \xmark & \cmark  & \cmark & \cmark & \xmark \\
        \bottomrule
      \end{tabular}
      }
      \vspace{-4mm}
    \end{table}


\section{Proposed Learning Framework}
\label{sec:frameworks}
\paragraph{Problem Formulation}
We now formulate the mapping from speech to semantics (IC/SL).
Consider some target SLU dataset $\mathcal{D} = \{\bm{A}^{(i)},\bm{W}^{(i)},\bm{S}^{(i)}, \bm{I}^{(i)}\}_{i=1}^{M}$ consisting of $M$ i.i.d. sequences, where $\bm{A}^{(i)},\bm{W}^{(i)},\bm{S}^{(i)}$ are the audio, word and slots sub-sequences respectively and $\bm{I}^{(i)}$ is their corresponding intent label. 
Note that $\bm{W}$ and $\bm{S}$ are sub-sequences of the same length, and $\bm{I}$ is a one hot vector.  
We are interested in finding the model $\theta_{SLU}^*$ where, 
    \begin{equation}
    \theta_{SLU}^* =  \underset{\theta}{\text{argmax}} \mathcal{L}_{SLU}(\theta_{SLU};\mathcal{D})
    = \underset{\theta}{\text{argmax}} 
    \mathop{\mathbb{E}_{(\bm{A},\bm{W},\bm{S},\bm{I})\sim\mathcal{D}}}
    \Big[ 
    \ln P(\bm{W},\bm{S},\bm{I} \mid \bm{A};\theta_{SLU}) 
    \Big]
    \end{equation}

At test time, an input audio sequence $\bm{a} = a_{1:T}$ and the sets of all possible word tokens $\mathcal{W}$, slots $\mathcal{S}$, and intents $\mathcal{I}$ are given.
We are then interested in decoding for its target word sequence $\bm{w}^* = w_{1:N}$, its slots sequence $\bm{s}^* = s_{1:N}$, and its intent label $\bm{i}^*$, where N is the number of word/slots tokens.
In the following subsections, we will describe an end-to-end implementation of our framework and its two baseline variants, depending on how $\theta_{SLU}$ is formalized\footnote{We abuse some notations by representing models by their model parameters, e.g. $\theta_{ASR}$ for the ASR model and $\theta_{BERT}$ for BERT.}.

    \subsection{End-to-End: Joint E2E ASR and BERT Fine-Tuning.}
    \label{subsec:end_to_end}
        To implement the end-to-end SLU model $\theta_{SLU}$, we take a pretrained E2E ASR and a pretrained deep contextualized LM, such as BERT, and jointly predict $\bm{W}$, $\bm{S}$ and $\bm{I}$ on $\mathcal{D}$.
        In this case, the pretraining objectives are ASR subword prediction for the ASR~\citep{lugosch2019speech,tomashenko2019recent} and Masked Language Modeling (MLM) for BERT. 
        During fine-tuning on $\mathcal{D}$, outputs from the ASR and BERT are concatenated to predict $\bm{S}$ and $\bm{I}$ with loss $\mathcal{L}_{NLU}$, while $\bm{W}$ is predicted with loss $\mathcal{L}_{ASR}$.
        The main benefit this formulation brings is that now $\bm{S}$ and $\bm{I}$ do not solely depend on an ASR top-1 hypothesis $\bm{W}^*$ during training, and the end-to-end fine-tuning objective is thus,
        \begin{equation}
        \mathcal{L}_{SLU}(\theta_{SLU};\mathcal{D}) = \mathcal{L}_{ASR}(\theta_{SLU};\mathcal{D}) + \mathcal{L}_{NLU}(\theta_{SLU};\mathcal{D}).
        \end{equation}

\begin{figure}[!htbp]
        \centering
        \includegraphics[width=.45\linewidth]{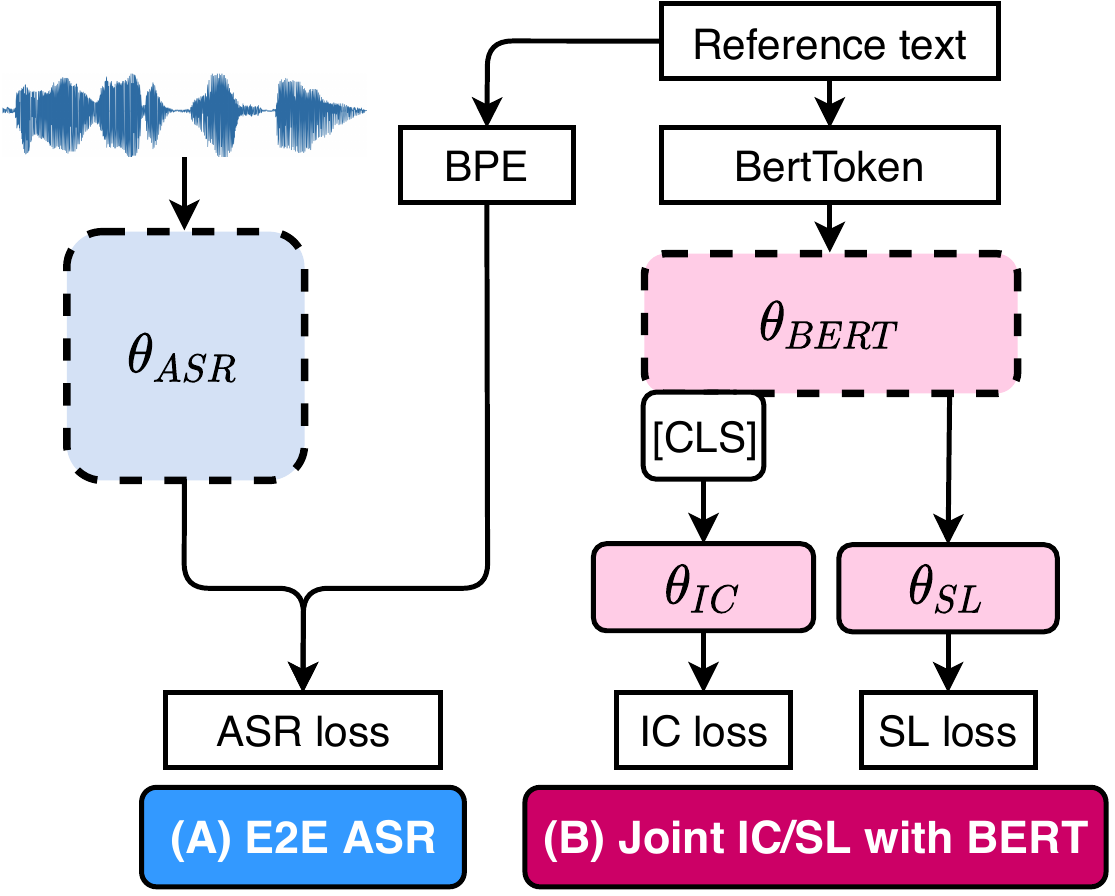}
        \caption{
            E2E ASR and BERT. 
            Note that $\theta_{ASR}$ and $\theta_{BERT}$ have different subword tokenizations: SentencePiece (BPE)~\citep{kudo2018sentencepiece} and BertToken.
            Dotted shapes are pretrained.
        }
        \label{fig:building_blocks_simple}
        \vspace{-3mm}
        \end{figure}

        \paragraph{Model Building Blocks: E2E ASR and BERT}
        A visualization of the model building blocks is in Figure~\ref{fig:building_blocks_simple}, where \underline{$\theta_{SLU} = \{\theta_{ASR}, \theta_{BERT}, \theta_{IC}, \theta_{SL}\}$}. 
        $\theta_{ASR}$ is the model parameter for the E2E ASR. 
        The choice of E2E ASR over hybrid ASR here is due to later on, the whole SLU model can backprop the errors from $\bm{S}$ and $\bm{I}$ through $\bm{A}$.
        The ASR objective $\mathcal{L}_{ASR}$ is formulated to maximize sequence-level log-likelihood, 
        \begin{equation}
        \mathcal{L}_{ASR}(\theta_{SLU};\mathcal{D}) = \mathcal{L}_{ASR}(\theta_{ASR};\mathcal{D}) = 
        \mathop{\mathbb{E}_{(\bm{A},\bm{W})\sim\mathcal{D}}}
        \Big[
        \ln P(\bm{W}\mid \bm{A}; \theta_{ASR}) 
        \Big]
        \end{equation}

        \paragraph{}Contextualized LM plays a critical role in the context of semantics understanding, and in this work, we opted to use BERT~\citep{devlin2018bert}, $\theta_{BERT}$, as the basis for \textit{jointly} predicting $\bm{S}$ and $\bm{I}$. 
        Following~\citep{chen2019bert}, $\bm{S}$ is predicted via an additional CRF/linear layer on top of BERT, and $\bm{I}$ is predicted on top of the BERT output of the [CLS] token.
        The additional model parameters for predicting SL and IC are $\theta_{SL}$ and $\theta_{IC}$, respectively. 
        Before writing down $\mathcal{L}_{NLU}$, we describe a masking operation because ASR and BERT typically employ different subword tokenization methods\footnote{Alternatively, it may be possible to pretrain ASR and BERT with the same tokenization method; yet, this implies that one can not use the pretrained models already available.}. 

        \vspace{-2mm}

        \newcommand{\colormatA}[1]{
        \stackrel{#1}{
        \vcenter{\hbox{ 
        \begin{tikzpicture}[scale=0.3, align=center]
        \filldraw[step=1,black,thin] (0.0, 0.0) grid (3, 5);
        \filldraw[shift={(0,0)}, fill=blue!700!white, draw=black, fill opacity=0.5] (0, 0) rectangle (3,1);
        \filldraw[shift={(0,2)}, fill=blue!700!white, draw=black, fill opacity=0.5] (0, 0) rectangle (3,1);
        \filldraw[shift={(0,4)}, fill=blue!700!white, draw=black, fill opacity=0.5] (0, 0) rectangle (3,1);
        \end{tikzpicture}
        }}}}
        
        \newcommand{\colormatB}[1]{
        \stackrel{#1}{
        \vcenter{\hbox{ 
        \begin{tikzpicture}[scale=0.3, align=center]
        \filldraw[step=1,black,thin] (0.0, 0.0) grid (3, 4);
        \filldraw[shift={(0,0)}, fill=red!700!white, draw=black, fill opacity=0.5] (0, 0) rectangle (3,1);
        \filldraw[shift={(0,1)}, fill=red!700!white, draw=black, fill opacity=0.5] (0, 0) rectangle (3,1);
        \filldraw[shift={(0,3)}, fill=red!700!white, draw=black, fill opacity=0.5] (0, 0) rectangle (3,1);
        \end{tikzpicture}
        }}}}
        
        \newcommand{\colormatC}[1]{
        \stackrel{#1}{
        \vcenter{\hbox{ 
        \begin{tikzpicture}[scale=0.3, align=center]
        \filldraw[step=1,black,thin] (0.0, 0.0) grid (10, 3);
        \filldraw[shift={(0,0)}, fill=blue!700!white, draw=black, fill opacity=0.5] (0, 0) rectangle (4,3);
        \filldraw[shift={(4,0)}, fill=red!700!white, draw=black, fill opacity=0.5] (0, 0) rectangle (6,3);
        \end{tikzpicture}
        }}}}
        \vspace{-3mm}
        \begin{equation}
        \label{eq:matrix_vis}
        concat\Big((\colormatA{M^{a}})^T \cdot \colormat{4}{5}{0}{0}{0}{0}{H^{a}} + (\colormatB{M^{b}})^T \cdot \colormat{6}{4}{0}{0}{0}{0}{H^{b}}
        \Big) = \colormatC{H^{cat}}
        \end{equation}
        
        \vspace{-2mm}
        \paragraph{Differentiate Through Subword Tokenizations}
        To concatenate $\theta_{ASR}$ and $\theta_{BERT}$ outputs along the hidden dimension, we need to make sure they have the same length along the token dimension\footnote{Here, we opted for the more straightforward operation possible. There are other more sophisticated solutions, such as attention mechanisms to align the two outputs, or gumbel softmax to backprop through the tokenizations.}. 
        We stored the first indices where $\bm{W}$ are broken down into subword tokens into a matrix: $M^{a}\in{\rm I\!R}^{N^{a}\times N}$ for $\theta_{ASR}$ and $M^{b}\in{\rm I\!R}^{N^{b}\times N}$ for $\theta_{BERT}$, 
        where $N$ be the number of tokens for $\bm{W}$ and $\bm{S}$, $N^{a}$ be the number of ASR subword tokens, and $N^{b}$ for BERT. 
        Let $H^{a}$ be the $\theta_{ASR}$ output matrix before softmax, and similarly $H^{b}$ for $\theta_{BERT}$. 
        The concatenated matrix $H^{cat}\in{\rm I\!R}^{N\times(512+768)}$ is given as $H^{cat} = \text{concat}(\left[ (M^{a})^T H^{a}, (M^{b})^T H^{b} \right], \text{dim=1})$, 
        where $512$ and $768$ are hidden dimensions for $\theta_{ASR}$ and $\theta_{BERT}$.
        A visualization of this process is Eq.~\ref{eq:matrix_vis}.
        We are now ready to describe $\mathcal{L}_{NLU}$: 
        \begin{align}
        \label{eq:e2E_nlu}
        \mathcal{L}_{NLU}(\theta_{SLU};\mathcal{D}) &= 
        \mathop{\mathbb{E}_{}}
        \Big[
        \ln P(\bm{S}\mid H^{cat};\theta_{SL}) + 
        \ln P(\bm{I}\mid H^{cat};\theta_{IC}),
        \Big] 
        \end{align}
        where sum of cross entropy losses for IC and SL are maximized, and $\theta_{ASR}$ and $\theta_{BERT}$ are updated through $H^{cat}$.
        Note here that ground truth $\bm{W}$ is used instead of $\bm{W}^*$ due to teacher forcing.
        
        \paragraph{Inference}
        Having obtained $\theta_{SLU}^*$ and given an audio sequence $\bm{a}$, the decoding procedure is,
        \begin{equation}
        \bm{w}^* = 
        \underset{w_{n}\in\mathcal{W}}{\text{argmax}}\prod_{n=1}^N p(w_{n}\mid w_{n-1:n-e}, \bm{a}; \theta_{SLU}^*), 
        \label{eq:e2e_inf_w}
        \end{equation}
        \begin{equation}
        \bm{i}^*,\bm{s}^* = 
        \underset{i\in\mathcal{I}}{\text{argmax }} p(i\mid \bm{w}^*, \bm{a}; \theta_{SLU}^*),
        \underset{s_{n}\in\mathcal{S}}{\text{argmax}}\prod_{n=1}^N p(s_{n}\mid \bm{w}^*, \bm{a}; \theta_{SLU}^*)
        \label{eq:e2e_inf_s}
        \end{equation}
        This two step decoding procedure, first $\bm{w}^*$ then $(\bm{i}^*,\bm{s}^*)$ is necessary for our framework even if it is trained end-to-end, given that no explicit serialization on $\bm{W}$ and $\bm{S}$ are imposed, like~\citep{tomashenko2019recent,haghani2018audio}. 
        $\bm{w}^*$ decoding has $w_{n-1:n-e}$ since there can be an optional $(e+1)$-gram LM, though we did not find it helpful and omitted it in the experiments; 
        while decoding for $(\bm{i}^*,\bm{s}^*)$, additional input $\bm{a}$ is given and we have $\bm{w}^*$ given the context from self-attention in BERT. 
        Note that here and throughout the work, we only take top-1 hypothesis $\bm{w}^*$ (instead of top-N) to decode for $(\bm{i}^*,\bm{s}^*)$.

    \subsection{Baselines}
    \label{subsec:speechbert}
    
        Two slight variations, 2-stage and SpeechBERT, for constructing $\theta_{SLU}$ are presented (refer to Figure~\ref{fig:overview} for illustration).
        They will be the baselines for the end-to-end approach. 
        \vspace{-2mm}
        \subsubsection{2-Stage: Cascading ASR Outputs to BERT}
        \vspace{-1mm}
        A natural baseline to the end-to-end approach is \textit{separately} pretrain and fine-tune $\theta_{ASR}$ and $\theta_{BERT}$, and during inference, cascade the top-1 ASR hypothesis $\bm{W}^*$ as input to BERT. 

        \vspace{-1mm}
        \subsubsection{SpeechBERT: BERT in Joint Speech-Text Embedding Space}
\begin{figure}[]
        \centering
        \includegraphics[width=0.9\linewidth]{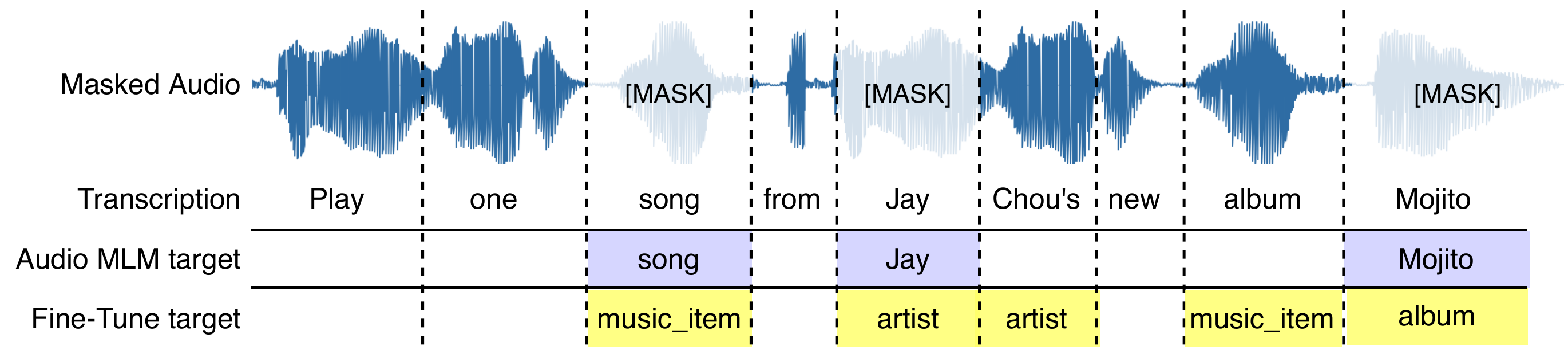}
        \caption{
        Illustration of SpeechBERT Audio MLM and IC/SL fine-tuning setup. 
        }
        \label{fig:audio_mlm}
        \vspace{-4mm}
        \end{figure}

        \vspace{-1mm}
        Another sensible way to construct $\theta_{SLU}$ is to somehow "adapt" the BERT model such that it can take audio has inputs and outputs IC/SL, while not compromising its original semantics learning capacity. 
        SpeechBERT~\citep{chuang2019speechbert} was initially proposed for Spoken Question Answering (SQA), but we found the core idea of training BERT with audio-text pairs fitting as another baseline for our end-to-end approach. 
        Three steps are involved in predicting semantics from speech with SpeechBERT.
        First, align audio segments to word tokens with a segmentation function $F_{seg}(\bm{a}): w_{n}\leftrightarrow\bm{a}_{u:v}$, where a word token $w_{n}$ is mapped to an audio segment $\bm{a}_{u:v}$ of the audio sequence $\bm{a}$.
        Although there is a line of work on unsupervised audio-text alignment, for example~\citep{chung2018unsupervised,kamper2019truly}, we opted to use force alignment as $F_{seg}$. 
        The quality of $F_{seg}$'s audio segment boundaries is vital, as it creates the input/output pairs for SpeechBERT's pretraining and fine-tuning.  
        Figure~\ref{fig:audio_mlm} illustrates the audio-text and audio-IC/SL pairs for SpeechBERT. 
        
        \vspace{-3mm}
        \paragraph{Pretraining: Mapping Audio Segments to Text}
        SpeechBERT is pretrained with Audio MLM on a separate dataset where paired audio-text pairs are available but not semantics labels. 
        Audio MLM is similar to MLM in BERT~\citep{devlin2018bert}, but with audio segments $\bm{a}_{u:v}$ as input and word token $w_{n}$ as the target. 
        Similar to MLM's dynamic masking policy~\citep{devlin2018bert}, parts of the audio segments are masked during training.
        An audio segment summarizer is needed to produce a single vector to represent variable-length audio segments, and following~\citep{chuang2019speechbert}, we implemented it with an encoder-decoder LSTM. 
        This pretraining step gradually adapts a pretrained BERT to a phonetic-semantic joint embedding space. 
        As before, we define $\theta_{SLU} = \{\theta_{ASR}, \theta_{BERT}, \theta_{IC}, \theta_{SL}\}$. 
        Unlike the end-to-end approach, though, $\theta_{ASR}$ is kept frozen throughout the SpeechBERT pretraining and fine-tuning phases.  
        Finally, note that as in Figure~\ref{fig:overview}, we make $\bm{a}$ to be the last hidden output from $\theta_{ASR}$ as opposed to MFCCs, as the original work was aimed at single speaker SQA.
        
        \vspace{-2mm}

        \vspace{-1mm}
        \paragraph{Fine-tuning: Mapping Audio Segments to IC/SL}
        The fine-tuning step is similar to Eq.~\ref{eq:e2E_nlu}, but $\theta_{ASR}^*$ is frozen and $F_{seg}$ and $\bm{W}$ are needed to align audio segments to their IC/SL:
        \begin{multline}
        \mathcal{L}_{NLU}(\theta_{SLU};\mathcal{D}) =
        \mathop{\mathbb{E}_{}}
        \Big[
        \ln P(\bm{S}\mid \bm{A},\bm{W}, F_{seg};\theta_{ASR}^*, \theta_{BERT}, \theta_{SL})\\
        + \ln P(\bm{I}\mid \bm{A},\bm{W}, F_{seg};\theta_{ASR}^*, \theta_{BERT}, \theta_{IC}),
        \Big] 
        \end{multline}
        
\vspace{-8mm}
\section{Experimental Setup}
\label{sec:exp}

    \vspace{-2mm}
    \paragraph{Datasets}
    Experiments are done on ATIS and SNIPS since their recordings are considerably smaller than those in-house SLU data used in~\citep{rao2020speech,coucke2018snips,haghani2018audio,serdyuk2018towards}. 
    ATIS~\citep{hemphill1990atis} contains 8hr of audio recordings of people making flight reservations with corresponding human transcripts. 
    A total of 5.2k utterances with more than 600 speakers are present. 
    SNIPS is another popular dataset (10.5hr), and given that the original training audio data was not released~\citep{coucke2018snips}, we used a commercial TTS service\footnote{Synthesized audios can potentially make pronunciation errors on proper nouns or technical terms that are probably out-of-vocabulary, inducing another error source in SLU modeling.} to synthesize audio from text data, similar to~\citep{huang2020learning}. 
    Different from~\citep{huang2020learning}, we synthesized SNIPS audio with 15 speakers, which we refer to as SNIPS-Multi\footnote{\url{https://github.com/aws-samples/aws-lex-noisy-spoken-language-understanding}}.
    Audios in ATIS and SNIPS-Multi are sampled at 16kHz.
    For the unlabeled data, we selected Librispeech 960hr (LS-960)~\citep{panayotov2015librispeech} and MS-SNSD (2.8hr)~\citep{reddy2019scalable}. 
    Besides the clean ATIS and SNIPS-Multi, models are evaluated on their noisy partition (augmented with MS-SNSD). 
    We made sure the noisy train and test splits in MS-SNSD do not overlap. 
    Lastly, while transcriptions are provided in ATIS and SNIPS, they are not normalized for ASR. 
    Text normalization is applied with an open-source software\footnote{\url{https://github.com/EFord36/normalise}}. 
    For ATIS, utterances are ignored if they contain words with multiple slot labels~\citep{tur2010left}. 
    The full details of our dataset statistics are in Appendix~\ref{app:corpora}.
    
    \vspace{-4mm}
    \paragraph{Hyperparameters and Compute Budget}
    All speech is represented as sequences of 83-dimensional Mel-scale filter bank with pitch, computed every 10ms.
    Global mean normalization is applied. 
    E2E ASR is implemented in ESPnet~\citep{watanabe2018espnet}, where it has 12 Transformer encoder layers and 6 decoder layers.
    The choice of the Transformer architecture~\citep{vaswani2017attention} is due to its empirical successes in~\citep{karita2019comparative} and concurrent SLU work~\citep{radfar2020end}.
    The E2E ASR is trained with hybrid CTC/attention loss~\citep{watanabe2017hybrid} (CTC weight is 0.3, attention weight is 0.7) with label smoothing.
    During ASR decoding, the beam size is set to 5 throughout this work, with scores from CTC, attention decoder, and an RNN-LM. 
    SpecAugment~\citep{park2019specaugment} is used by default for data augmentation.  
    SentencePiece (BPE) vocabulary size is set to 1k for ATIS and SNIPS-Multi.
    Model is optimized with noam~\citep{vaswani2017attention} and trained until convergence. 
    BERT is a bert-base-uncased from HuggingFace~\citep{wolf2019huggingface}. 
    All experiments were done on a single Nvidia V100. 
    Training took a few hours to complete for our ASR and SLU models. 
    For NLU (jointBERT~\citep{chen2019bert}), training takes a few minutes.
    
    \vspace{-3mm}
    \subsection{E2E Evaluation with Slots Edit $F_{1}$ score.}
    \vspace{-2mm}
    Our framework is evaluated with an end-to-end evaluation metric, termed the slots edit $F_{1}$. 
    Unlike slots $F_{1}$ score, slots edit $F_{1}$ accounts for instances where predicted sequences have different lengths as the ground truth.  
    To calculate the score, we first aligned the predicted text and oracle text.
    For each slot label $v\in\mathcal{V}$, where $\mathcal{V}$ is the set of all possible slot labels except for the "O" tag, we calculate the \underline{insertion} (false positive, FP), \underline{deletion} (false negative, FN), and \underline{substitution} (FN and FP) of its slots value. 
    Slots edit $F_{1}$ is the harmonic mean of precision and recall over all slots:
    \vspace{0mm}
    \begin{equation}
    \label{eq:slot_edit_f1}
    \text{slots edit } F_{1} = \frac{\sum_{v\in\mathcal{V}} 2\times \text{TP}_{v}}
    {
    \sum_{v\in\mathcal{V}}\Big[ (2\times \text{TP}_{v}) + \text{FP}_{v} + \text{FN}_{v}\Big]
    }
    \end{equation} 
    We notice that there were only two prior works evaluated their models with E2E evaluation criteria~\citep{rao2020speech,haghani2018audio}.
    Although slots edit $F_{1}$ is not perfect\footnote{Slots edit $F_{1}$ score double-penalizes \underline{substitution} errors, or English phrases that contains multiple words.}, we encourage readers to look at Table~\ref{table:edit_f1_illustration} in the Appendix for why these E2E evaluations are needed.
    
    \vspace{-3mm}
    \subsection{Two-Stage Fine-tuning}
    \label{subsec:e2e_fine_tune}
    \vspace{-2mm}
    The default training method for our end-to-end approach is to first pretrain the ASR component on LS-960 (noted as $\mathcal{{\widetilde{D}}}$) before fine-tuning the whole model $\theta_{SLU}$ on the target SLU corpus $\mathcal{D}$.
    An observation from the experiment was that ASR is harder than IC/SL. See Figure~\ref{fig:plot_e2e_1_stage_atis_acc} in the Appendix for reference, where IC/SL losses converge much faster than ASR loss.
    Therefore, alternatively, we train the end-to-end approach in two-stage: 
    pretrain ASR, then fine-tune ASR on $\mathcal{D}$, and finally jointly fine-tune for ASR and IC/SL on $\mathcal{D}$:
    $\mathcal{L}_{ASR}(\theta_{ASR};\mathcal{{\widetilde{D}}})
    \longrightarrow 
    \mathcal{L}_{ASR}(\theta_{ASR};\mathcal{D})
    \longrightarrow 
    \mathcal{L}_{SLU}(\theta_{SLU};\mathcal{D}).$
    
    \vspace{-3mm}
    \subsection{Main Results on Clean and Noisy SLU}
    \vspace{-2mm}
    We benchmarked our proposed framework with several prior works on ATIS and SNIPS, and Table~\ref{tab:main_results} presents their WER, slots edit F1 and intent F1 results. 
    All experimental results were averaged over at least three trials with random seeds.
    JointBERT~\citep{chen2019bert} is our NLU baseline, where BERT is jointly fine-tuned for IC/SL, and it gets around 95\% slots edit $F_{1}$ and over 98\% IC F1. 
    Since JointBERT has access to the oracle text, this is the upper bound to our SLU models with speech as input. 
    CLM-BERT~\citep{caostyle} explored using in-house conversational LM for NLU.
    We replicated~\citep{tomashenko2019recent}, where an E2E ASR (Listen, Attend and Spell~\citep{chan2016listen}) directly predicts interleaving word and slots tokens (serialized output), and optimized with CTC over words and slots.
    We also experimented with replacing E2E ASR with a Kaldi hybrid ASR. 
    \vspace{-1mm}
    \paragraph{} Table~\ref{tab:main_results} presents results on clean test data.
    Both our proposed end-to-end and baselines approach surpassed prior SLU work in terms of WER, slots edit $F_{1}$, and intent F1. 
    We did not compare to E2E SLU work if SL is not modeled, see Table~\ref{tab:comparison}.
    We hypothesize the performance gain originates from our choices of (1) adopting pretrained E2E ASR and self-supervised LM like BERT, (2) applying text-norm on target transcriptions for training the ASR, and (3) joint fine-tuning text and IC/SL. 
    \vspace{-3mm}
    \paragraph{} To quantify model robustness under noisy settings, we augmented ATIS and SNIP-Multi with environmental noise from MS-SNSD, which is a common scenario where users utter their spoken commands. 
    The incentive here is how 'noise' was abused in some SLU literature, where ASR errors were treated as the noise source instead of modeling error, see~\citep{huang2020learning,wangasr}.
    Results on noisy test reveal that those work well on ATIS or SNIPS may break under realistic noises. 
    Although our models are trained with SpecAugment~\citep{park2019specaugment}, there is still a 5\% and 10\% relative drop on ATIS and SNIPS for the E2E approach, and a 4-27\% drops for the baselines. 
    This consequence directly motivated Section~\ref{subsec:nosie_fine_tune}.

    \vspace{-1mm}
    \begin{table}[!htbp]
    \vspace{-3mm}
    \caption{
    WER, slots edit $F_{1}$ and intent $F_{1}$ on ATIS and SNIPS-Multi (clean test). 
    Models use Librispeech 960h (LS-960) and MS-SNSD as additional unlabeled training data. 
    ATIS and SNIPS-Multi are augmented with real environmental noises (noisy test) to evaluate model noise-robustness. 
    We compared our proposed end-to-end and baseline approaches with prior SLU work and the NLU counterpart, where oracle text is assumed.
    Results indicate that our semi-supervised framework is effective in data scarcity setting, exceeding prior work in WER and IC/SL while approaching the NLU upper bound.
    }
    \vspace{2mm}
    \label{tab:main_results}
    \centering 
    \resizebox{1.0\textwidth}{!}{
    \begin{tabular}{lccccccc}
    \toprule
    \multirow{2}{*}{Frameworks} & Unlabelled & \multicolumn{3}{c}{clean test} & \multicolumn{3}{c}{noisy test} \\
    \cmidrule(lr){3-5}\cmidrule(lr){6-8}
    {} & {Semantics Data} & {WER} & {slots edit $F_{1}$} & {intent $F_{1}$} & {WER} & {slots edit $F_{1}$} & {intent $F_{1}$} \\
    \midrule
    \midrule
    \multicolumn{8}{l}{\textbf{ATIS with Oracle Text}}\\
    JointBERT~\citep{chen2019bert} & & - & 95.64 & 98.99  & - & - & - \\
    
    \midrule
    \multicolumn{8}{l}{\textbf{Proposed E2E on ATIS}}\\
    End-to-End w/ two-stage fine-tune     & LS-960    & 2.18  & \bf{95.88} & 97.26 & 9.62 & \bf{91.54} & \bf{96.14} \\
    
    \midrule
    \multicolumn{8}{l}{\textbf{Proposed Baseline on ATIS}}\\
    2-Stage Baseline                       & LS-960     & 1.38 & 93.69     & 97.01     & 8.98 & 90.09 & 95.74 \\
    SpeechBERT Baseline                    & LS-960    & 1.4  & 92.36 & \bf{97.4} & 9.0 & 81.72 & 94.05 \\

    \midrule 
    \multicolumn{8}{l}{\textbf{Prior Work on ATIS}}\\
    ASR-Robust Embed~\citep{huang2020learning} & WSJ & 15.55 & -        & 95.65     & -    & - & - \\
    Kaldi Hybrid ASR+BERT                   & LS-960     & 13.31 & 85.13    & 94.56     & 44.72 & 69.55 & 88.94 \\
    ASR+CLM-BERT~\citep{caostyle}           & in-house   & 18.4. & 93.8\footnotemark & 97.1  & - & - & - \\
    LAS+CTC~\citep{tomashenko2019recent}    & LS-460   & 8.32  & 86.85 & - & - & - & - \\
    
    \midrule
    \midrule

    \multicolumn{8}{l}{\textbf{SNIPS with Oracle Text}}\\
    JointBERT~\citep{chen2019bert} & & - & 94.71 & 98.43  & - & - & - \\

    \midrule
    \multicolumn{8}{l}{\textbf{Proposed E2E on SNIPS}}\\
    End-to-End  w/ two-stage fine-tune   & LS-960      & 11.86 & \bf{83.41} & \bf{98.65} & 20.9 & \bf{74.22} & 95.90 \\
    
    \midrule
    \multicolumn{8}{l}{\textbf{Proposed Baseline on SNIPS}}\\
    2-Stage Baseline                & LS-960    & 11.87 & 81.51     & 98.18 & 21.2 & 72.39 & \bf{95.59} \\
    
    \midrule 
    \multicolumn{8}{l}{\textbf{Prior Work on SNIPS}}\\
    ASR-Robust Embed~\citep{huang2020learning} & WSJ & 45.56    & - & 89.55 & - & - & - \\
    Kaldi Hybrid ASR+BERT        & LS-960    & 30.89 & 68.35     & 94.76 & 52.28 & 49.46 & 76.98 \\
    ASR+CLM-BERT~\citep{caostyle}        & in-house  & 16.2  & 89.3  & 98.6 & - & - & - \\

    \bottomrule
    \end{tabular}
    }
    \end{table}
    \addtocounter{footnote}{0}
    \footnotetext{For ASR+CLM-BERT~\citep{caostyle}, model predictions are evaluated only if its ASR hypothesis and human transcription have the same number of tokens. }

    \vspace{-2mm}
    \subsection{Environmental Noise Augmentation}
    \label{subsec:nosie_fine_tune}
    \vspace{-2mm}
    To further improve the noise-robustness of our learning framework, we augment our framework training with MS-SNSD.  
    Although there is much work on E2E speech enhancement~\citep{subramanian2019speech}, we found that merely augmenting the training data with a diverse set of environmental noises works well.
    We followed the noise augmentation protocol described in~\citep{reddy2019scalable}, where for each training sample, five noise files are randomly sampled and added to the clean file with SNR levels of $[0, 10, 20, 30, 40]$dB, resulting in a five-fold data augmentation.
    Table~\ref{tab:noise_results} shows our proposed models trained with noise augmentation. We first observe that compared to Table~\ref{tab:main_results}, now there is a minimal performance drop when these noises are present (noisy test).
    On ATIS, our E2E approach reaches 95.46\% for SL and 97.4\% for IC, which is merely a 1-2\% drop from the clean test data. 
    Compared to models trained without noises, there is a 4\% SL improvement over its clean counterpart, and almost 40\% improvement over Kaldi's hybrid ASR. 
    We also observe that on ATIS, the E2E model now performs on par with the NLU model with oracle text as input.
    On SNIPS, a similar trend is observed, despite there is still a large gap between our SLU models and their NLU upper bound. 
    The 13\% WER could explain the performance gap on SNIP (c.f. 2\% on ATIS), which motivated our next modification. 

    
    \vspace{-2mm}
    \begin{table}[!htbp]
    \caption{
    Noise augmentation reduces model degradation when environmental noises are present. 
    }
    \vspace{2mm}
    \label{tab:noise_results}
    \centering 
    \resizebox{1\textwidth}{!}{
    \begin{tabular}{lccccccc}
    \toprule
    \multirow{2}{*}{Frameworks} & \multicolumn{3}{c}{clean test} && \multicolumn{3}{c}{noisy test} \\
    \cmidrule(lr){2-4}\cmidrule(lr){6-8}
    {} & {WER} & {slots edit $F_{1}$} & {intent $F_{1}$} && {WER} & {slots edit $F_{1}$} & {intent $F_{1}$} \\
    \midrule
    \midrule
    \multicolumn{8}{l}{\textbf{ATIS with Oracle Text}}\\
    JointBERT~\citep{chen2019bert} & - & 95.64 & 98.99  && - & - & - \\

    \midrule
    \multicolumn{8}{l}{\textbf{Proposed on ATIS w/ Noise Aug.}}\\
    End-to-End w/ two-stage fine-tune      & 2.13  & \bf{96.38} & \bf{97.65} && 3.6 & \bf{95.46} & \bf{97.40} \\
    2-Stage Baseline                & 1.73  & 93.41 & 96.79     && 3.5 & 92.52 & 96.49 \\
    SpeechBERT Baseline             & 1.8   & 92.66 & 96.91     && 3.6 & 88.7 & 96.15 \\

    \midrule
    \midrule

    \multicolumn{8}{l}{\textbf{SNIPS with Oracle Text}}\\
    JointBERT~\citep{chen2019bert} & - & 94.71 & 98.43  && - & - & - \\
    
    \midrule
    \multicolumn{8}{l}{\textbf{Proposed on SNIPS w/ Noise Aug.}}\\
    End-to-End w/ two-stage fine-tune  & 13.5  & \bf{82.12} & \bf{98.28} && 15.3 & \bf{80.02} & \bf{97.90} \\ 
    2-Stage Baseline        & 13.37 & 79.65 & 97.82 && 15.23 & 77.58 & 97.59 \\

    \bottomrule
    \end{tabular}}
    \vspace{-4mm}
    \end{table}
    
    \subsection{Recovering Domain-Specific Words via Knowledge-Base (KB) Refinement}
    \vspace{-2mm}    
    Another observation in our experiments was that many domain-specific words are hard to predict \textit{perfectly} even for humans.  
    For example, in SNIPS, there is an extensive list of artists and album names.
    A refinement step is further added after text $\bm{w}^*$ and slot $\bm{s}^*$ sequences are decoded to "correct" the wrongly decoded words by replacing them with the closest matched words from the target corpus.
    First, for each slot $s^*$, we construct a knowledge-base $\text{KB}_{s^*}$ that contains all words $s^*$ matched in $\mathcal{D}$.  
    Then, for each predicted pair ($w^{*}$, $s^*$), $w^{*}$ is replaced with $w^{*}_{r}$ from $\text{KB}_{s^*}$ that has the highest embedding similarity with $w^{*}$.
    Embeddings are retrieved from a pretrained BERT. 
    Succinctly, 

    \begin{equation}
    \label{eq:slot_edit_f1}
    (w^{*}, s^*)\longrightarrow
    (w^{*}_{r} = \underset{m\in\text{KB}_{s^*}}{\text{argmax }} dot\Big(\text{BERT}(w_{*}),  \text{BERT}(m)\Big), s^*)
    \end{equation}
    \vspace{-4mm}
    \paragraph{}Table~\ref{tab:KB_results} shows the effectiveness of KB refinement on SNIPS, where although the WER remained the same, slots edit $F_{1}$ greatly improved. 
    Our E2E approach now reaches 90\% on SL, less than 5\% from the NLU upper bound and around a 9\% improvement over the E2E baseline. 
    Theoretically, we could have an iterative refinement process and potentially reach even higher $F_{1}$ scores. 
    \vspace{-2mm}
    
    \begin{table}[!htbp]
    \vspace{-2mm}
    \caption{
    KB refinement "correct" decoded text where many domain-specific entities are present. 
    }
    \vspace{2mm}
    \label{tab:KB_results}
    \centering 
    \resizebox{1\textwidth}{!}{
    \begin{tabular}{lccccccc}
    \toprule
    \multirow{2}{*}{Frameworks} & \multicolumn{3}{c}{clean test} && \multicolumn{3}{c}{noisy test} \\
    \cmidrule(lr){2-4}\cmidrule(lr){6-8}
    {} & {WER} & {slots edit $F_{1}$} & {intent $F_{1}$} && {WER} & {slots edit $F_{1}$} & {intent $F_{1}$} \\
    \midrule
    \midrule

    
    \multicolumn{8}{l}{\textbf{Proposed on SNIPS w/ KB refinement}}\\
    End-to-End w/ two-stage fine-tune      & 11.86 & 83.41 & \bf{98.65} && 20.9 & 74.22 & 95.90 \\
    \qquad + KB refinement          &       & \bf{90.86} &       &&      & 81.96 &  \\
    \quad + noise augmentation      & 13.5 & 82.12 & 98.28 && 15.3 & 80.02 & \bf{97.90} \\ 
    \qquad + KB refinement          &     & 89.46 &       &&      & \bf{87.51} &     \\

    \bottomrule
    \end{tabular}
    }
    \vspace{-4mm}
    \end{table}

\section{Conclusions and Future Work}
\label{sec:conclusion}
\vspace{-1mm}
This work attempts to respond to a classic paper "What is left to be understood in ATIS?~\citep{tur2010left}", and to the advancement put forward by contextualized LM and end-to-end methods up against semantics understanding. 
We proposed a learning framework that works well under data scarcity and noisy settings while re-examining the current paradigm in terms of how SLU is modeled and evaluated. 
We compared our semi-supervised methods against prior work quantitatively with a new E2E evaluation metric, the slots edit $F_{1}$ score, on two public SLU corpora. 
\vspace{-1mm}
\paragraph{} We showed for the first time that an SLU model with speech as input could perform on par with NLU models on ATIS, entering the 5\% "corpus error/ambiguities" range noted in~\citep{tur2010left,bechet2018atis}. 
However, have we solved the task once and for all? 
Referencing the SNIPS results, the answer is a resounding no. 
Unsolved questions remain, such as the prospect of building a single framework for \textbf{multi-lingual} SLU~\citep{glass1995multilingual}, or the need for a more spontaneous SLU corpus that is not limited to short segments of spoken commands. 
For future work, we plan to relax the semi-supervised constraints with unsupervised speech representations~\citep{baevski2020wav2vec}. 


\section*{Broader Impact}
\vspace{-3mm}
In this work, we have shown that in the data scarcity regime, our semi-supervised frameworks can still achieve competitive performance as those where oracle text is present. 
Looking further ahead, we hope this will motivate a line of work on building multilingual SLU models with little to no transcriptions, making the SLU technology available to the 7,000 languages and dialects around the world. 

\begin{ack}
\vspace{-3mm}
The SpeechBERT experiments in this paper are run by Yung-Sung Chuang from National Taiwan University (NTU). We thank the regular advice from Hung-yi Lee from NTU. 
We thank Su Zhu from Shanghai Jiao Tong University, Alice Coucke from Sonos, Inc., and Chao-Wei Huang from NTU for the various spontaneous exchanges with us. 
We thank Nanxin Chen from Johns Hopkins University, Erica Cooper from National Institute of Informatics, and Alexander H. Liu, Wei Fang, Fan-Keng Sun, and Jim Glass from MIT for their comments on the paper presentation.
We also thank the anonymous reviewers for their comments. 
\end{ack}

\bibliography{main}

\bibliographystyle{abbrvnat}

\newpage

\appendix
\renewcommand{\thesection}{\Alph{section}}

\section*{Appendices}


\section{Formulating Slot Labeling as Intent Detection -- Why is it Problematic?}
\label{app:problem_with_FSC}
A recent trend in end-to-end SLU is to formulate it solely as an intent detection problem. 
The idea is simple, slot labels with all their possible slot values combinations are flattened into one-hot vectors as the classification objective. 
Therefore, the optimization objective of the network is cross-entropy loss, as classification is easier than regression in most cases. 
This trend was likely brought up by the release of the new Fluent Speech Corpus (FSC), as that was how the corpus was designed~\citep{lugosch2019speech}. 
Nevertheless, this is not scalable! 
Imagine if there are 10 slot labels and each slot has 1000 slot values (this is not ridiculous as a working SLU model should handle as many users queries as possible), there are $1000^{10}$ possible combinations if slot labeling is formulated as intent classification!
For example, the slot label 'destination\_city' could have slot values 'New York City,' 'Boston,' 'San Francisco,' 'Toronto,' etc. It can take on slots values of any cities in the world. 
Therefore, one of the design notes we kept in mind in coming up with these frameworks is to have both IC and SL as the output target sequences. 

\section{Full Summary Table of Previous Work}
\label{app:summary_table}

    \begin{table}[htbp]
      \centering
      \caption{Full comparison of our proposed framework with prior work in terms of modeling and evaluation.
      In addition to previous work on E2E SLU, we included here some NLU work with oracle text as input.}
      \resizebox{\columnwidth}{!}{
      \begin{tabular}{lccccccc}
        \toprule
        \multirow{2}{*}{Model}  &  \multicolumn{5}{c}{Modeling}  &  \multicolumn{2}{c}{Evaluation}  \\
        \cmidrule(lr){2-6}\cmidrule(lr){7-8}
           &  Input  &  Output  & Noise/Error & Semi-Supervised & E2E & E2E evaluation & Noise Robustness \\
        \midrule
        \multicolumn{6}{l}{\textbf{Proposed}}\\
        2-Stage     & speech  &  text, intent, slots  &  \cmark & \cmark &  \xmark & \cmark & \cmark \\
        End-to-End  & speech  &  text, intent, slots  &  \cmark & \cmark &  \cmark & \cmark & \cmark \\
        SpeechBERT  & speech  &  text, intent, slots  &  \cmark & \cmark &  \xmark & \cmark & \cmark \\
        \midrule 

        \multicolumn{6}{l}{\textbf{Prior Work}}\\
        \citep{serdyuk2018towards}$^{*}$                 &  speech   &  intent only          & \cmark & \xmark  & \cmark & \xmark & \cmark \\
        \citep{lugosch2019speech,wang2020large,cho2020speech}$^{*}$    &  speech   &  intent only          & \xmark & \cmark  & \cmark & \xmark & \xmark \\
        \citep{radfar2020end}$^{*}$                      &  speech   &  intent only          & \xmark & \xmark  & \cmark & \xmark & \xmark \\
        \citep{ghannay2018end}                      &  speech   & text, intent, slots   & \cmark & \cmark  & \cmark & \xmark & \xmark \\
        \citep{haghani2018audio}                    & speech    & text, intent, slots   & \xmark & \xmark  & \cmark & \xmark & \xmark \\
        \citep{tomashenko2019recent}                & speech    & text, intent, slots   & \xmark & \cmark  & \cmark & \xmark & \xmark \\
        \citep{huang2020learning}                   & text      & intent only           & \cmark & \xmark  & \xmark & \xmark & \xmark \\
        \citep{chen2019bert}                        & text      & intent, slots         & \xmark & \cmark  & \xmark & \xmark & \xmark \\
        \citep{zhu2017encoder}                      & text      & intent, slots         & \xmark & \xmark  & \xmark & \xmark & \xmark \\
        \bottomrule
      \end{tabular}
      }
    \end{table}

\newpage
\section{Model Architectures}
\label{app:model_architectures}
        
        \begin{figure}[!htbp]
        \centering
        \includegraphics[width=1.0\linewidth]{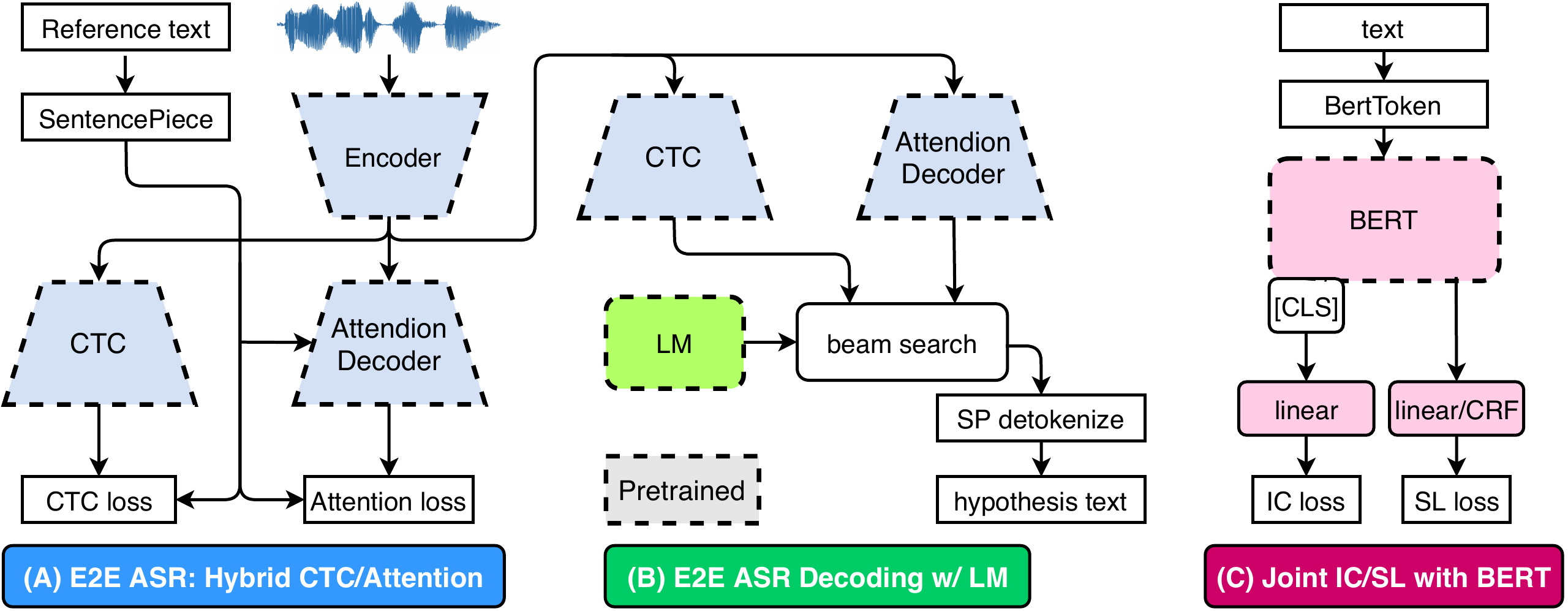}
        \caption{
            Basic building blocks: E2E ASR $\theta_{ASR}$ and fine-tuned BERT $\theta_{BERT}$. 
            (A) illustrates E2E ASR with hybrid CTC/Attention losses. 
            (B) shows beam search decoding for E2E ASR with scores from CTC, Attention decoder and LM.
            (C) shows fine-tuned BERT with joint IC and SL losses. Intent is predicted on top of the [CLS] token. 
            Note that $\theta_{ASR}$ and $\theta_{BERT}$ have different subword tokenizations: SentencePiece (BPE)~\citep{kudo2018sentencepiece} and BertToken.
            Shapes in dotted lines are pretrained.
        }
        \label{fig:building_blocks}
        \vspace{-3mm}
        \end{figure}

\newpage 
\section{Illustrated Example: How does Slots Edit $F_{1}$ Score fit in?}
\label{app:edit_f1_illustration}
    
    \begin{table}[!htbp]
    \caption{
    Examples to illustrate why an end-to-end IC/SL evaluation protocol is needed and how it is computed. 
    (A.1) shows a sentence with perfect text recognition does not capture any semantics.
    (A.2) shows that despite a sentence that has high WER, semantics are captured! 
    (A.3) shows the issue of evaluating with \textit{only} slots $F_{1}$: although slots types are correct, their values are not.\\
    (B) A word-slot pair is shown as a \word{word\bw{[tag]}}.
    \ignore{Words highlighted in grey} are ignored in evaluation since they are labeled as 'O.'
    Highlighted words: \rwtwo{insertion}, \del{deletion}, \sub{substitution} and \correct{correct}, signified the word-level differences per slot between the ground-truth and predicted sequences after alignment during evaluation.
    \rwtwo{Insertion} is counted as FP, \del{deletion} as FN, \sub{substitution} as both FP and FN, \correct{correct} as TP. 
    Slots edit $F_{1}$ is calculated according to Eq.~\ref{eq:slot_edit_f1}.
    The table is best when viewed in color.
    }
    \begin{center}
    \footnotesize
    \begin{tabularx}{\linewidth}{L}
        \toprule
        \subtitle{(A) Ground Truth} The \word{potato\bw{[food]}} and \word{cauliflower\bw{[food]}} are both in season to make \word{combo\bw{[food]}} \word{breads\bw{[food]}}, \word{mounds\bw{[food]}}, or \word{pads\bw{[food]}}. $\Longrightarrow$\textbf{intent: baking}
    
        \subtitle{(A.1) Sample Prediction} The \word{\del{potato}\bw{[food]}} \word{\rwtwo{potato}\bw{[sports]}} and \word{\del{cauliflower}\bw{[food]}} \word{\rwtwo{cauliflower}\bw{[sports]}} are both in season to make \word{\del{combo}\bw{[food]}} \word{\rwtwo{combo}\bw{[sports]}} \word{\del{breads}\bw{[food]}} \word{\rwtwo{breads}\bw{[sports]}}, \word{\del{mounds}\bw{[food]}} \word{\rwtwo{mounds}\bw{[sports]}}, or \word{\del{pads}\bw{[food]}} \word{\rwtwo{pads}\bw{[sports]}}. $\Longrightarrow$\textbf{intent: \del{baking} \rwtwo{sports}}
        
        \subtitle{(A.2) Sample Prediction} \ignore{blaw} \word{potato\bw{[food]}} \ignore{blaw} \word{cauliflower\bw{[food]}} \ignore{blaw} \ignore{blaw} \ignore{blaw} \ignore{blaw} \ignore{blaw} \ignore{blaw} \word{combo\bw{[food]}} \word{breads\bw{[food]}}, \word{mounds\bw{[food]}}, \ignore{blaw} \word{pads\bw{[food]}}. $\Longrightarrow$\textbf{intent: \correct{baking}}
        
        \subtitle{(A.3) Sample Prediction} The \word{\sub{tomato}\bw{[food]}} and \word{\sub{cabbage}\bw{[food]}} are both in season to make \word{\sub{sour}\bw{[food]}} \word{\correct{breads}\bw{[food]}}, \word{\sub{mud}\bw{[food]}}, or \word{\sub{pets}\bw{[food]}}. $\Longrightarrow$\textbf{intent: \correct{baking}}
    
        \\ \midrule
        
        \subtitle{(B) Ground Truth} please find a flight \word{round\bw{[B-roundtrip]}} \word{trip\bw{[I-roundtrip]}} from \word{los\bw{[B-fromloc.city]}} \word{angeles\bw{[I-fromloc.city]}} to \word{tacoma\bw{[B-toloc.city]}} \word{washington\bw{[B-toloc.state]}} with a stopover in \word{san\bw{[B-stoploc.city]}} \word{francisco\bw{[I-stoploc.city]}} \word{not\bw{[B-cost.relative]}} \word{exceeding\bw{[I-cost.relative]}} the price of \word{three\bw{[B-fare]}} \word{hundred\bw{[I-fare]}} \word{dollars\bw{[I-fare]}} for \word{june\bw{[B-depart.month]}} \word{tenth\bw{[B-depart.day]}} \word{nineteen\bw{[B-depart.year]}} \word{ninety\bw{[I-depart.year]}} \word{three\bw{[I-depart.year]}} $\Longrightarrow$\textbf{intent: flight}
        
        \subtitle{(B) Sample Prediction} \ignore{*} find a \ignore{flights} \word{\correct{round}\bw{[B-roundtrip]}} \word{\correct{trip}\bw{[I-roundtrip]}} from \word{\correct{los}\bw{[B-fromloc.city]}} \word{\correct{angeles}\bw{[I-fromloc.city]}} to \word{\del{tacoma}\bw{[B-toloc.city]}} \word{\rwtwo{taco}\bw{[B-toloc.city]}} \word{\rwtwo{ma}\bw{[I-toloc.city]}} \word{\correct{washington}\bw{[B-toloc.state]}} with a stopover in \word{\correct{san}\bw{[B-stoploc.city]}} \word{\del{francisco}\bw{[I-stoploc.city]}} \word{\rwtwo{francisco}\bw{[I-toloc.city]}} \word{\correct{not}\bw{[B-cost.relative]}} \word{\sub{exciting}\bw{[I-cost.relative]}} the price of \word{\correct{three}\bw{[B-fare]}} \word{\correct{hundred}\bw{[I-fare]}} \word{\sub{dollar}\bw{[I-fare]}} for \word{\correct{june}\bw{[B-depart.month]}} \word{\correct{tenth}\bw{[B-depart.day]}} \word{\correct{nineteen}\bw{[B-depart.year]}} \word{\sub{nineteen}\bw{[I-depart.year]}} \word{\correct{three}\bw{[I-depart.year]}} $\Longrightarrow$\textbf{intent: \correct{flight}}
        
        \\ \bottomrule
    
    \end{tabularx}
    \end{center}
    \label{table:edit_f1_illustration}
    \end{table}

\newpage
\section{Dataset Statistics}
\label{app:corpora}
The detailed statistics of the corpora used in this work are presented. ATIS and SNIPS are standard SLU corpora. However, given that the SNIPS training data is not released to the public (Section 2.1 of~\citep{coucke2018snips}), SNIPS audios are synthesized with a commercial TTS service. The SNIPS audios were synthesized with 15 speakers\footnote{Speaker lists: Aditi, Amy, Brian, Emma, Geraint, Ivy, Joey, Justin, Kendra, Kevin, Kimberly, Matthew, Nicole, Raveena, Russell, Salli.} and is referred to as SNIPS-Multi here. We also randomly selected a single speaker, Emma, as SNIPS-Single, and evaluated our framework on that. The noise corpus and noise augmentation procedure is based on MS-SNSD~\citep{reddy2019scalable}, where nine types\footnote{MS-SNSD noise types: vacuum cleaner, typing, copy machine, shutting door, neighbor speaking, munching, babble, announcement, air conditioner.} of environmental noises are present. For noise augmentation, we made sure that train, valid, and test partition \textbf{do not} have overlapping noise files. ATIS-Noise is ATIS augmented with MS-SNSD, and SNIPS-Multi-Noise is SNIPS-Multi augmented with MS-SNSD. Lastly, LibriSpeech 960 (LS-960) is used for the pretraining, only paralleled audio-text data $\mathcal{{\widetilde{D}}}$ is required in our work. 

\paragraph{Why did we not choose MUSAN for noise augmentation?}
MUSAN~\citep{snyder2015musan}is a widely-adopted speech, music, and noise corpora now widely incorporated for training SOTA speaker recognition system~\citep{snyder2018x}, and it does meet our purpose here. 
Our original goal was to focus on the noise robustness for SLU, where potential secondary background speakers may be present. To design a model that does direct noise suppression, we had in mind a controllable corpus that is designed for speech enhancement, like MS-SNSD. 

\paragraph{SNIPS-Multi has 160 hours of data. That is a lot!} 160 hours of audio data is indeed a lot; however, it is not that much compared to what some of the previous work used in their settings, see, for example, ~\citep{coucke2018snips}.
Besides, keep in mind that SNIPS-Multi has the same content as SNIPS-Single. 
The only difference between them is speaker variability, which should be learned to be ignored by the model (see section 3.2 Speaker Adaptive Training (SAT) in~\citep{tomashenko2019recent}). 
Therefore, the same mistake that was made in SNIPS-Single supposedly would likely re-occur in SNIPS-Multi.

    \label{tab:corpora}
    \centering
    \begin{tabular}{llccccc}
    \toprule
    Type & Corpus & Train & Valid & Test & Speakers & Unique Transcriptions \\
    \midrule
    $\mathcal{D}$      & ATIS          &  8 hr     &  1 hr     &  1.5hr   &  678  &  5.2k \\
    $\mathcal{D}$      & SNIPS-Single  &  10.5 hr  &  35 min   &  35 min  &  1    &  14.5k  \\
    $\mathcal{D}$      & SNIPS-Multi   &  160  hr  &  8.5 hr   &  8.5hr  &  15   &  14.5k  \\
    $\mathcal{D_{N}}$  & MS-SNSD       &  2.8 hr  &  30 min   & 40 min  &  -    &  -  \\
    $\mathcal{D}+\mathcal{D_{N}}$      & ATIS-Noise     &  50 hr  &  5 hr   & 7hr   &  678  &  5.2k \\
    $\mathcal{D}+\mathcal{D_{N}}$      & SNIPS-Multi-Noise  &  800 hr  &  42.5 hr  &  42.5 hr  &  15 &  14.5k \\
    $\mathcal{{\widetilde{D}}}$        &  LS-960  &  960 hr  &  10 hr & 10 hr &  2338  &  -  \\
    \bottomrule
    \end{tabular}



    
    

\newpage
\section{Training plots}
\label{app:train_plots}

\begin{figure*}[!hbtp] 
	\centering
	\begin{subfigure}[h]{0.8\textwidth}
		\centering
		\includegraphics[width=\textwidth]{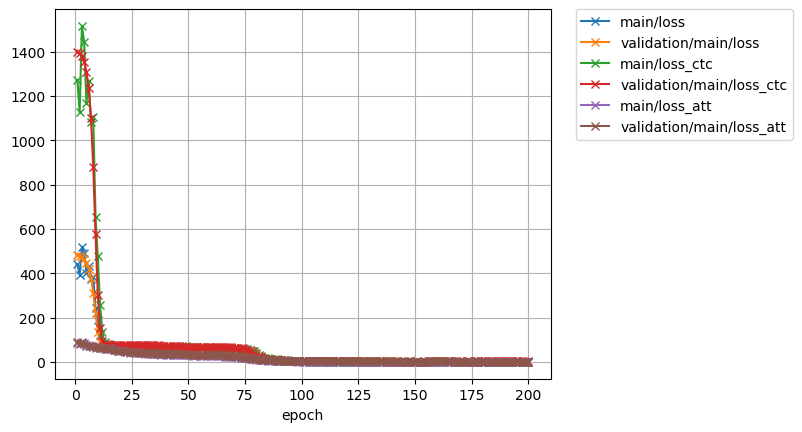}	
		\vspace{-0.38cm}
		\caption{Training Loss}
	\end{subfigure}
		\begin{subfigure}[h]{0.8\textwidth}
		\centering
		\includegraphics[width=\textwidth]{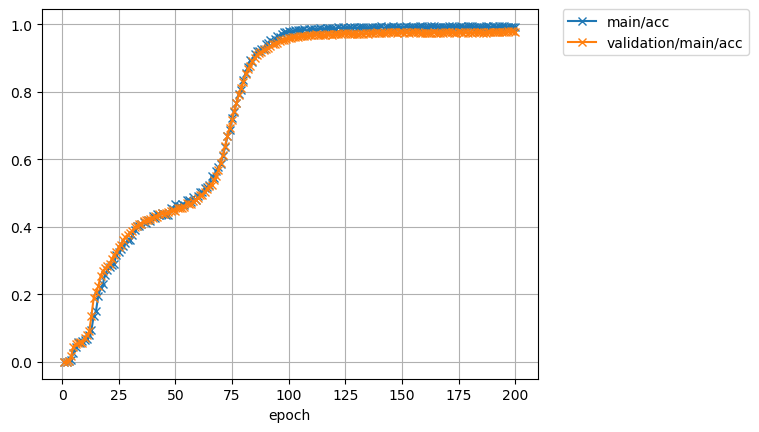}	
		\vspace{-0.38cm}
		\caption{Training Accuracy}
	\end{subfigure}
		\begin{subfigure}[h]{0.8\textwidth}
		\centering
		\includegraphics[width=\textwidth]{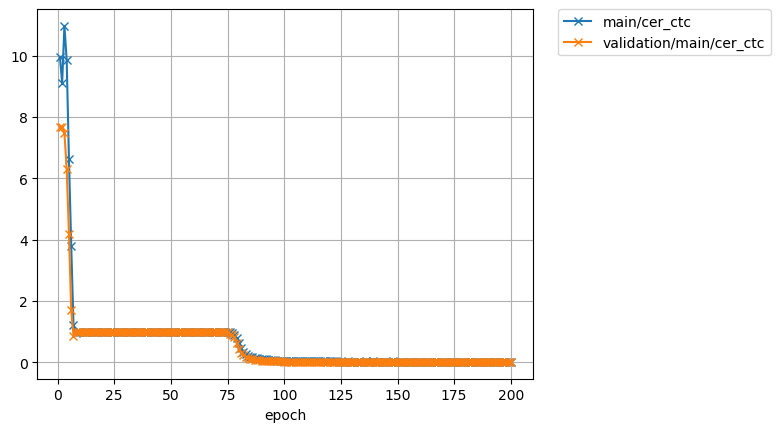}	
		\vspace{-0.38cm}
		\caption{Training CTC Character Error Rate (CER)}
	\end{subfigure}
	\caption{Plots for our E2E ASR training on ATIS.}
	\vspace{-0.2cm}
	\label{fig:plot_e2e_asr_atis}
\end{figure*}

\begin{figure*}[!t] 
	\centering
	\begin{subfigure}[h]{0.8\textwidth}
		\centering
		\includegraphics[width=\textwidth]{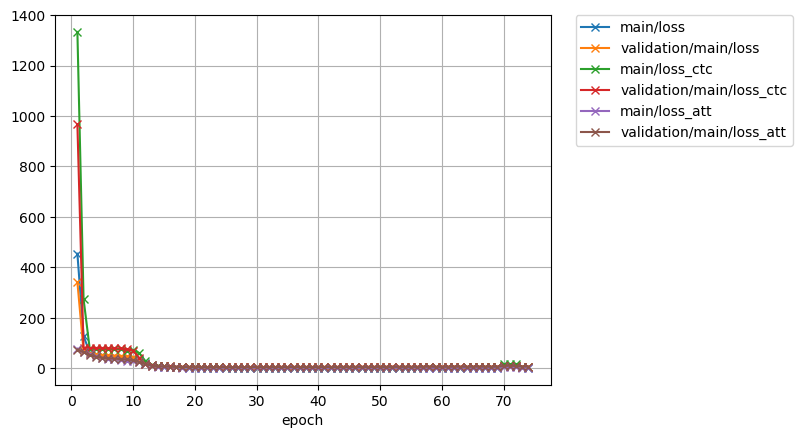}	
		\vspace{-0.38cm}
		\caption{Training Loss}
	\end{subfigure}
		\begin{subfigure}[h]{0.8\textwidth}
		\centering
		\includegraphics[width=\textwidth]{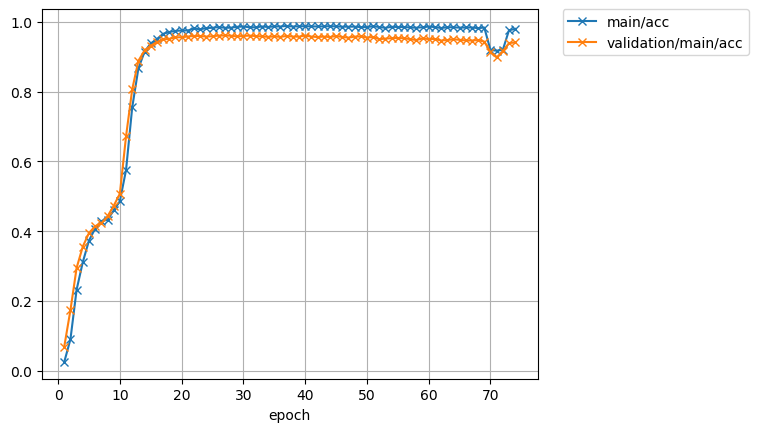}	
		\vspace{-0.38cm}
		\caption{Training Accuracy}
	\end{subfigure}
		\begin{subfigure}[h]{0.8\textwidth}
		\centering
		\includegraphics[width=\textwidth]{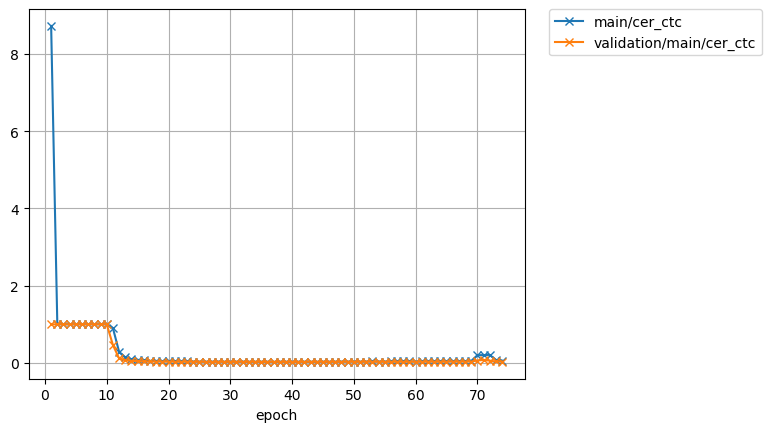}	
		\vspace{-0.38cm}
		\caption{Training CTC CER}
	\end{subfigure}
	\caption{Plots for our E2E ASR training with noise-augmentation on ATIS.}
	\vspace{-0.2cm}
	\label{fig:plot_e2e_asr_noise_atis}
\end{figure*}

\begin{figure*}[!t] 
	\centering
	\begin{subfigure}[h]{0.8\textwidth}
		\centering
		\includegraphics[width=\textwidth]{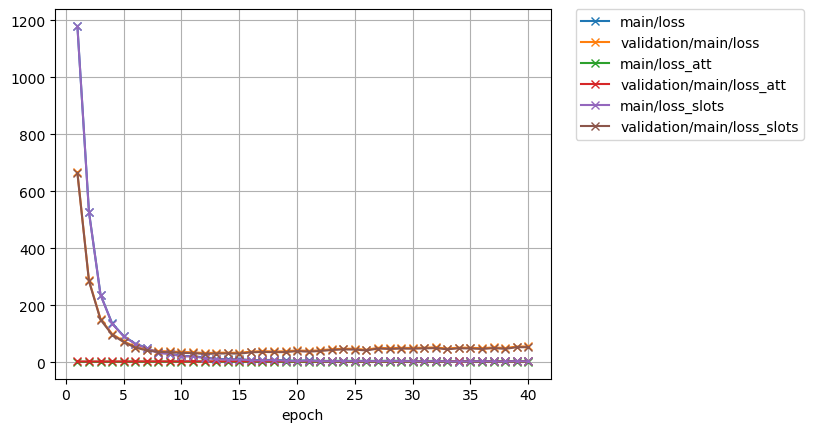}	
		\vspace{-0.38cm}
		\caption{Training Loss}
	\end{subfigure}
		\begin{subfigure}[h]{0.8\textwidth}
		\centering
		\includegraphics[width=\textwidth]{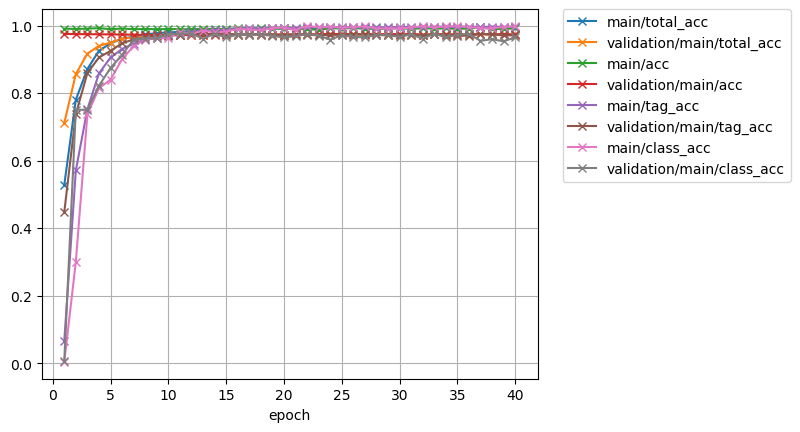}	
		\vspace{-0.38cm}
		\caption{Training Accuracy}
	\end{subfigure}
	\caption{Plots for our end-to-end approach \textit{with} two-stage fine-tuning on ATIS. 
	Note that here we fine-tuned without CTC loss, so there is not a CTC CER plot.}
	\vspace{-0.2cm}
	\label{fig:plot_e2e_2_stage_atis}
\end{figure*}

\begin{figure*}[!t] 
	\centering
	\begin{subfigure}[h]{0.8\textwidth}
		\centering
		\includegraphics[width=\textwidth]{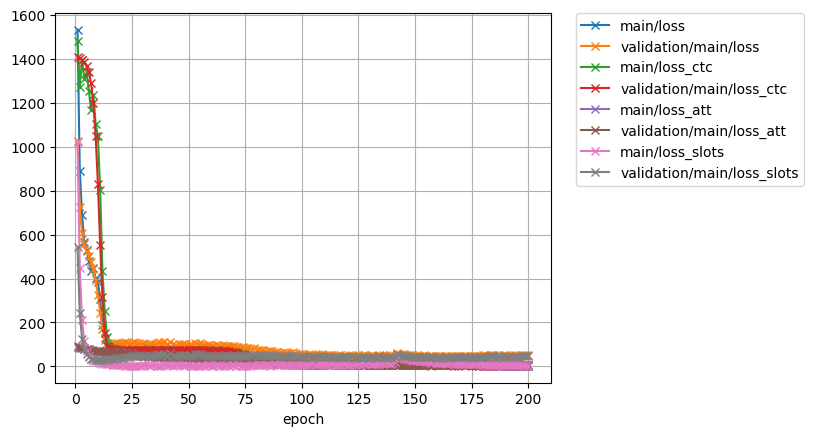}	
		\vspace{-0.38cm}
		\caption{Training Loss}
	\end{subfigure}
		\begin{subfigure}[h]{0.8\textwidth}
		\centering
		\includegraphics[width=\textwidth]{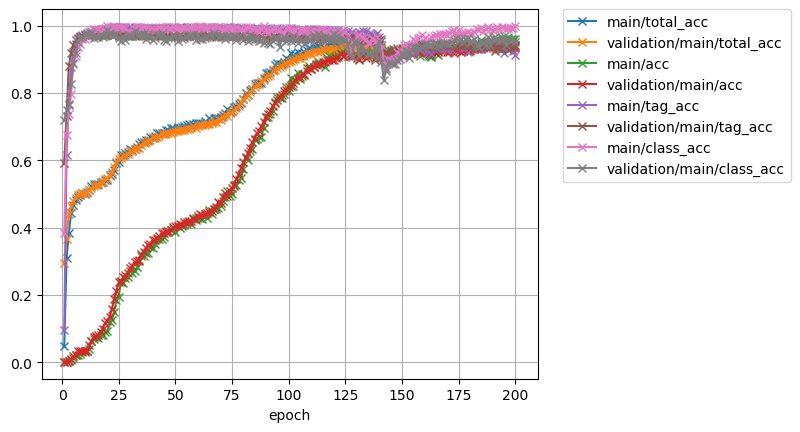}	
		\vspace{-0.38cm}
		\caption{Training Accuracy}
		\label{fig:plot_e2e_1_stage_atis_acc}
	\end{subfigure}
		\begin{subfigure}[h]{0.8\textwidth}
		\centering
		\includegraphics[width=\textwidth]{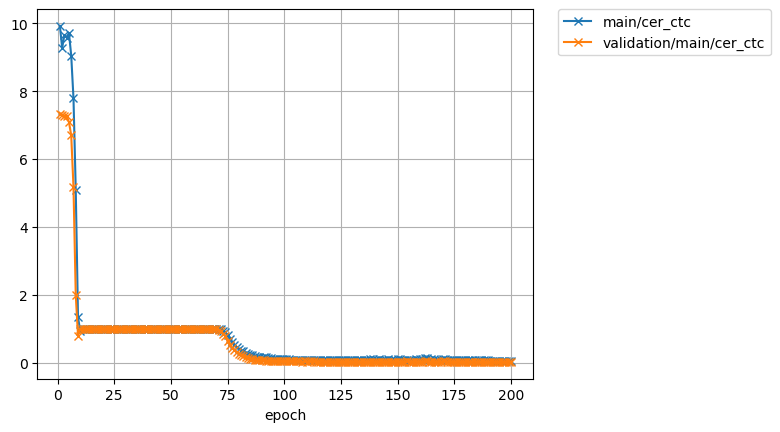}	
		\vspace{-0.38cm}
		\caption{Training CTC CER}
	\end{subfigure}
	\caption{Plots for our end-to-end approach \textit{without} two-stage fine-tuning on ATIS. 
	Note the curves in~\ref{fig:plot_e2e_1_stage_atis_acc} suggests that ASR is a much harder task than IC/SL.}
	\vspace{-0.2cm}
	\label{fig:plot_e2e_1_stage_atis}
\end{figure*}

\end{document}